# Detecting facial landmarks in the video based on a hybrid framework

Nian Cai[1, *], Zhineng Lin[1], Fu Zhang[1], Guandong Cen[1], Han Wang[2]

[1] School of Information Engineering, Guangdong University of Technology, 510006 Guangzhou, China

[2] School of Electromechanical Engineering, Guangdong University of Technology, 510006 Guangzhou, China

Corresponding author: Nian Cai, email: cainian@gdut.edu.cn

## Abstract

To dynamically detect the facial landmarks in the video, we propose a novel hybrid framework termed as detection-tracking-detection (DTD). First, the face bounding box is achieved from the first frame of the video sequence based on a traditional face detection method. Then, a landmark detector detects the facial landmarks, which is based on a cascaded deep convolution neural network (DCNN). Next, the face bounding box in the current frame is estimated and validated after the facial landmarks in the previous frame are tracked based on the median flow. Finally, the facial landmarks in the current frame are exactly detected from the validated face bounding box via the landmark detector. Experimental results indicate that the proposed framework can detect the facial landmarks in the video sequence more effectively and with lower consuming time compared to the frame-by-frame method via the DCNN.

## 1 Introduction

Facial landmark detection refers to locate facial landmarks by supervised or semi-supervised methods starting from face bounding boxes achieved by the face detectors [29]. Facial landmarks (e.g. eye, nose and mouth) imply the facial semantic information, which can improve the performance of face recognition/verification [1, 3], expression analysis [4, 6] and pose estimation [6, 10], and be applied to facial animation [2]. There are many facial landmark detection methods studied in static "in the wild" images recently. These methods can be roughly divided into three categories: 1) the methods [9, 10] with classify sliding search window to perform facial landmark detection. However, it is possible for them to detect non-plausible facial landmarks and be vulnerable to situation of ambiguity or corruption since their assumptions are that the landmarks are independent of each other. 2) Active Shape Model (ASM) and Active Appearance Model (AAM) [11-13]. The models update the positions of facial landmarks iteratively with generative model to fit global facial appearance, which are robust to partial occlusions. However, both appearance initialization and sufficient iteration number are critical in the case of extreme head pose. 3) Regression-based methods [14-16]. They are more efficient to predict coordinates of facial landmarks directly from facial features, such as Haar-like features, SIFT, local binary patterns (LBP).

Among the regression-based methods, deep-learning-based methods perform more excellent performance than most of the state of the art methods in the field of facial landmark detection [5-8]. This is because deep learning can discover intrinsic data structure in the big data [27]. Sun et al. [5] proposed a three-level cascaded deep convolutional neural network (DCNN) to detect the facial landmarks in the images. Actually, the method is a coarse-to-fine manner for facial landmark detection. A coarse facial landmark detection is implemented at the first level of the network. Then, a fine detection process is performed by the following two levels of the network to achieve the final facial landmarks. Experimental results indicated that their method was robust than the traditional methods and some commercial systems. Zhou et al. [7] designed a four-level cascaded DCNN to detect 68 facial landmarks in the images. 51 inner points and 17 contour points are estimated by the cascaded DCNN and by the DCNN, respectively. The facial landmarks are determined by combining the estimated points. Experimental results indicated the validation of their method for extensive facial landmark detection. Zhang et al. [6] proposed a facial landmark detection method for the static images by deep multi-task learning. To robustly detect the facial landmarks, they put forward a novel tasks-constrained deep model, with task-wise early stopping to facilitate learning convergence. Experimental results indicated that their scheme could robustly detect the facial landmarks in the cases of severe occlusions and large pose variations compared to existing methods. Different from the above methods employing the CNNs to construct the deep models, Zhang et al. [8] cascaded a few successive stacked auto-encoder networks (SANs) to detect the facial landmarks in the images. A low-resolution version of the detected face is input into the first SAN to estimate a roughly accurate shape. And the following SANs progressively refine the landmark locations by means of the local features extracted around the current landmarks, which are the outputs of the previous SAN. Experimental results showed that their method outperformed the state-of-the-art methods.

Although the existing deep-learning-based methods can perform excellent performance in facial landmark detection, they are all applied to detect the landmarks from the labeled face bounding boxes in the static images. We can predict that these methods can also detect the facial landmarks in the videos via the frame-by-frame scheme if the face bounding boxes are labeled manually or detected by means of face detectors automatically. However, it is impossible to beforehand label the face bounding boxes from the frames of the real-world videos. So, face detection is an indispensable procedure for facial landmark detection in the real-world videos. It is well-known that face detection is an open and challenging problem since pose variations, occlusions or bad illuminations often occur in the video sequences. Therefore, the existing deep-learning-based facial landmark detection methods maybe fail to detect the facial landmarks in the videos if they cannot effectively achieve the face bounding boxes via face detection. Also, it will cost much time if we detect the face bounding box from the whole frame via the frame-by-frame scheme, especially when the video is high-resolution.

To dynamically detect the facial landmarks in the videos, we propose a novel hybrid framework to perform facial landmark detection in the videos. The proposed framework integrates a cascaded DCNN for facial landmark detection, the median flow method [17] for facial landmark tracking, and a traditional face detector [20] for local face detection. So, we call it as the detection-tracking-detection (DTD) framework for simplicity in this paper. To the best of our knowledge, this is also the first attempt to introduce the DCNN into the dynamic facial landmark detection in the videos. First, a traditional face detector is employed to detect the face bounding box from the first frame of the video sequence. Here, we term this face detection process as global face detection. Then, a cascade DCNN is designed to detect the facial landmarks in the first frame from the face bounding box. Next, the estimated facial landmarks in the current frame are achieved via the median flow method and then used to estimate the current face bounding box. To avoid the drift, the estimated face bounding box is validated by the traditional face detector. Here, we term the estimation and validation of the face bounding box in the current frame as local face detection. This procedure will reduce the consuming time compared to detecting it from the whole current frame. Finally, the facial landmarks in the current frame are exactly detected from the validated face bounding box by means of the cascade DCNN.

## 2 Proposed Method

The proposed DTD hybrid framework includes four main procedures, which are global face detection, facial landmark detection, facial landmark tracking and local face detection. As illustrated in Fig. 1, global face detection is implemented by a traditional face detector, which detects the face bounding box from the whole image of the first frame of the video. Then, the facial landmarks in the first frame are detected via a cascaded DCNN. For the next frame, the facial landmarks in the current frame are estimated from those achieved in the previous frame via the median flow method, which is a promising tracking method. Next, local face detection is implemented to achieve the validated face bounding box in the current frame. This procedure includes two sub-procedures, which are face bounding box estimation and face bounding box validation. The face bounding box in the current frame is estimated from the estimated facial landmarks by means of the tracking method. Then, the estimated face bounding box is validated the traditional face detector in [20]. The validated face bounding box achieved by local face detection is used to detect the facial landmarks in the current frame based on the cascaded DCNN. The facial landmarks can be dynamically detected in the video when all the frames of the video undergo the procedures mentioned above.

Section 2.1 describes the network architecture of a cascaded DCNN in detail, which is used to implement the task of facial landmark detection. The scheme of facial landmark tracking is introduced in Section 2.2. And global/local face detection is introduced in Section 2.3.

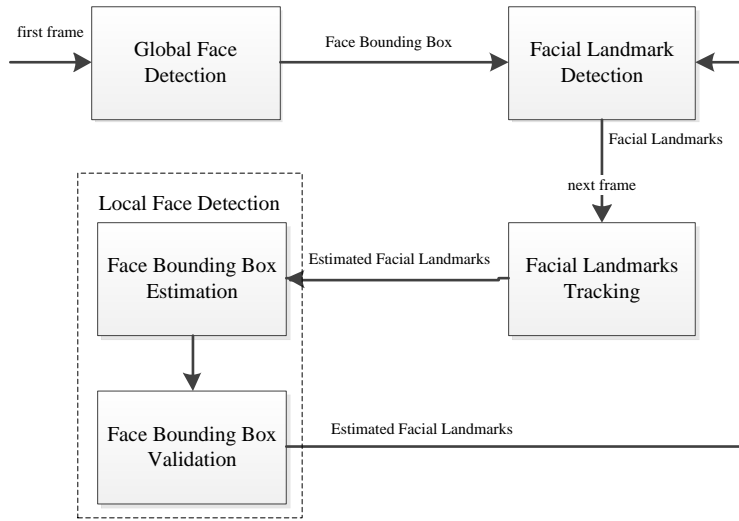

Fig. 1 The DTD framework for the facial landmarks in the video sequence

### 2.1 Network architecture of a cascaded DCNN

In this paper, we design a three-level cascaded DCNN for facial landmark detection in different face regions, whose network structure is the same as that in [5]. However, different from [5], we use the Rectified Linear Unit

(ReLU) [23] for nonlinear transform after the convolutional layers. The ReLU is a nonlinearity mapping $\sigma(x) = \max(x, 0)$ which makes the convergence fast while still achieving good performance.

As shown in Table 1, Level 1 of the cascaded DCNN has three sub-networks denoted by F1, EN1 and NM1. F1 sub-network uses the whole face as the input to predict all the landmarks. The eyes-and-nose image patch covering the eyes and the nose is used as the input of EN1 sub-network to predict two eye centers and one nose tip, which is the top and middle part of the face. NM1 sub-network uses the nose-and-mouth patch covering the nose and the mouth to predict the nose tip and two mouth corners, which is the middle and bottom part of the face. Each of Level 2 and Level 3 employs a common shallower structure since their tasks are relatively easy, in which two convolution neural networks (CNNs) are employed to predict each facial landmark. Thus, there are 10 CNNs in each of Level 2 and Level 3, whose inputs are the corresponding patches as illustrated in Table 1.

The deep structure of the DCNN in Level 1 contains four convolutional layers, three max-pooling layers and two fully connected layers. The shallower structure in Level 2 and Level 3 includes three convolutional layers, two max-pooling layers and one fully connected layer.

Table 1. The inputs of all the levels of the cascaded DCNN

| cascaded DCNN | net | input patches | labels |
|---|---|---|---|
| level 1 | F1 | whole face | all landmarks |
|  | EN1 | top and middle of face | eyes + nose |
|  | NM1 | middle and bottom of face | nose + mouth |
| level 2/level 3 | LE1/LE2 | left eye patches | left eye |
|  | RE1/RE2 | right eye patches | right eye |
|  | N1/N2 | nose patches | nose |
|  | LM1/LM2 | left mouth patches | left mouth |
|  | RM1/RM2 | right mouth patches | right mouth |

2.2 Facial landmark tracking

It is a simple idea that we employ the above cascaded DCNN to detect the facial landmarks in the video by means of a frame-by-frame scheme. To implement the detection task, we must achieve the face bounding box before facial landmark detection via some face detector since the face bounding box is the input of the cascaded DCNN. Obviously, the cascaded DCNN does not function if the face bounding box is not detected. However, the fact that the face detector does not detect the face often occurs in the video due to occlusions, pose variations and bad illuminations. Here, we introduce the idea of object tracking into the facial landmark detection via the cascaded DCNN to solve the problem that the face bounding box is not well detected, even not detected.

Kalal et al. [17] proposed a facilitated object tracker called median flow method, which was based on Lucas-Kanade tracker [24]. The method assumes that a good tracker should be irrelevant to the direction of time-flow. It means that the backward tracking from time t+1 to t is effective if the forward tracking succeeds. Experimental results indicated that the method can greatly improve the tracking reliability. Inspired by [17], we incorporate the median flow into the proposed framework for facial landmark detection in the videos.

Only five facial landmarks detected by the cascaded DCNN are not enough for stable tracking. In order to improve the effectiveness of tracking, 15 simulated points around each facial landmark are achieved in a grid rule. That is, 80 points including the simulated points and five facial landmarks are used for tracking via the median flow.

The original 80 points $x_t$ in the previous frame are forward tracked by the pyramidal Lucas-Kanade tracker and estimate the corresponding points $x_{t+1}$ in the current frame. This is a forward trajectory. The estimated points $x_{t+1}$ are backward tracked to the previous frame and estimate the points $\hat{x}_t$. This is a forward trajectory. Thus, a Forward-Backward error occurs between the original points and the estimated points, which is denoted by the distance $d = \|x_t - \hat{x}_t\|_2^2$. The not-well-estimated points in the current frame are the estimated points in the current frame tracked by the original points in the previous frame which satisfies the condition that the distance between the estimated points $\hat{x}_t$ and the original points $x_{t-1}$ are greater than the median distance d/2. These not-well-estimated points are filled out. And the remaining estimated points are used to estimate the face bounding box in the current frame. Five facial landmarks in the current frame are estimated by the estimated bounding box. 80 new points in the current frame are generated by five estimated facial landmarks and used to track the facial landmarks in the next frame by means of the above tracking scheme.

2.3 Global/local face detection

Face detection is indispensable for facial landmark detection and has been extensively studied for decades. Viola-Jones [20] face detector is a promising and facilitated method among many face detectors and widely used by many researchers. So, we also employ this face detector in our proposed framework for facial landmark detection. The face detector is realized by classifying the sliding window. It indicates that the sizes of the sliding window and the image involving the face influence the time effectiveness of the detector. When the size of the sliding window is fixed, the more the size of the image is, the more consuming time the detector costs. If face detection is implemented on the whole frame when we detect the facial landmarks frame by frame, we can predict that it will cost much time. In this paper, we incorporate the idea of object tracking into facial landmark detection and put forward global/local face detection to reduce the consuming time. For the first frame, we detect the face bounding box from the whole frame via the Viola-Jones face detector. This operation is called global face detection. For the other frames, we first achieve the estimated face bounding box in the current frame by the tracked facial landmarks in the previous frame. Then, the estimated face bounding box is validated via the Viola-Jones face detector. The estimation and validation operations are called local face detection because we only detect the face from a local region.

# 3 Experimental results and discussions

## 3.1 Experimental settings

All the experiments are performed on an Intel Xeon E5-2630 v3 (2.40GHz) machine with the operating system Ubuntu, which has a 64GB RAM and a commercial graphics processing units (GPU) NVIDIA Quadro K4200 with 4GB memory. We use Caffe [29] to off-line train the cascaded DCNN for facial landmark detection.

## 3.2 Offline training the cascaded DCNN

### 3.2.1 Training set

10,000 static color images in [5] that involve the faces are employed to construct the training dataset for the cascaded DCNN, which are the images selected from the Labeled Face in the Wild (LFW) [28] and the images coming from the web. Each original image is rotated 5 degrees clockwise and 5 degrees counterclockwise, respectively. Then, each image with the clockwise or counterclockwise rotation is further mirror rotated. Also, each original image is mirror rotated. All the images above are labeled with the face bounding box and five facial landmarks. Next, the face subimage is extracted from each image according to the face bounding box. Finally, the face subimages are converted to grayscale subimages, whose pixel values are normalized to zero-mean and unit-variance and whose sizes are all resized to the size of the input layers of F1 sub-network. Thus, these grayscale subimages construct the training set of the cascaded DCNN for facial landmark detection, which includes 60,000 grayscale subimages.

For each subimage in the training set, at Level 1, three image patches, i.e. the whole face, the eyes-and-nose image patch and the nose-and-mouth image patch, are input into the corresponding DCNNs. At Level 2 and Level 3, we take two training patches centered at positions randomly shifted from the ground truth position of each facial landmark as the inputs of the corresponding CNNs. The coordinate values of each facial landmark are normalized in the range of [0, 1] according to corresponding patches.

### 3.2.2 Training procedure

At the training stage, the goal of the cascaded DCNN in Section 2.1 is to predict the facial landmarks which are close to the ground truth facial landmarks according to the inputting face bounding box. The training procedure is to estimate the network parameters. It can be implemented by minimizing the mean squared error (MSE) over a dataset. Stochastic gradient descent with standard back-propagation [26, 27] is employed to minimize MSE. At each iteration, mini-batch is randomly drawn to calculate the gradients approximately, which is of high efficiency for a large-scale dataset.

## 3.3 Experimental comparisons and analysis

The testing video in this paper is selected from the 300-VW test data [21] containing 114 videos and separated into three categories. It belongs to the third category, in which the video is captured in totally arbitrary conditions (including severe occlusions and extreme illuminations). The length of the video is 1'33", including 2798 frames with the size of 1280*720 pixels. For fair comparisons, we combine the facial landmark detection method in [5] with the traditional face detector in [20]. However, different from our proposed framework, the method combining [5] with [20] detect the facial landmarks in the video frame by frame. Figs. 2-5 illustrate the facial landmark detection results, which are achieved by our proposed DTD framework and by the method combining [5] with [20], respectively.

Two methods employ the same scheme to detect the facial landmarks in the first frame of the video based on the same training set. So, they achieve the same results as shown in Fig. 2. However, for the remaining frames, two methods employ different schemes to detect the facial landmarks. The method combing [5] with [20] is a frame-by-frame scheme. Our proposed framework is a dynamic hybrid scheme incorporating the tracking idea. So, different facial landmark detection results on the remaining frames are achieved by the two methods. The method combining [5] with [20] misdetect and even does not detect the facial landmarks in many frames, such as in the 682th, 1242th and 1673th frames as shown in Figs. 3(a)-5(a). This is because the face bounding box is maybe misdetected or even not detected when the face detector is employed to detect the face frame by frame. Our proposed framework, however, performs good face bounding box detection via the procedures of facial landmark tracking and face bounding box validation. Therefore, compared to the method combining [5] with [20], our proposed framework can more effectively detect the facial landmarks in the video in a dynamic scheme. There are a very few not-ideal landmarks existed in the results achieved by our proposed method. It is well-known that large-area occlusion and large pose variation are the open and challenging problems in the fields of computer vision.

To compare the performance in computational burden, we make a statistical analysis on the consuming time with which the methods detect the all frames. Sometimes several faces are misdetected by the method combining [5] with [20] although only one face exists in the video. So, for fair comparisons, we only count the consuming time costed by the method combining [5] with [20] from only one possible face subimage. The proposed DTD framework takes 39.93ms per frame while the method combining [5] with [20] takes 133.77ms per frame. Obviously, the proposed DTD framework has excellent real-time performance compared to the method combining [5] with [20] detecting the face bounding box from the whole frame. This is because local face detection in our framework costs less consuming time than global face detection.

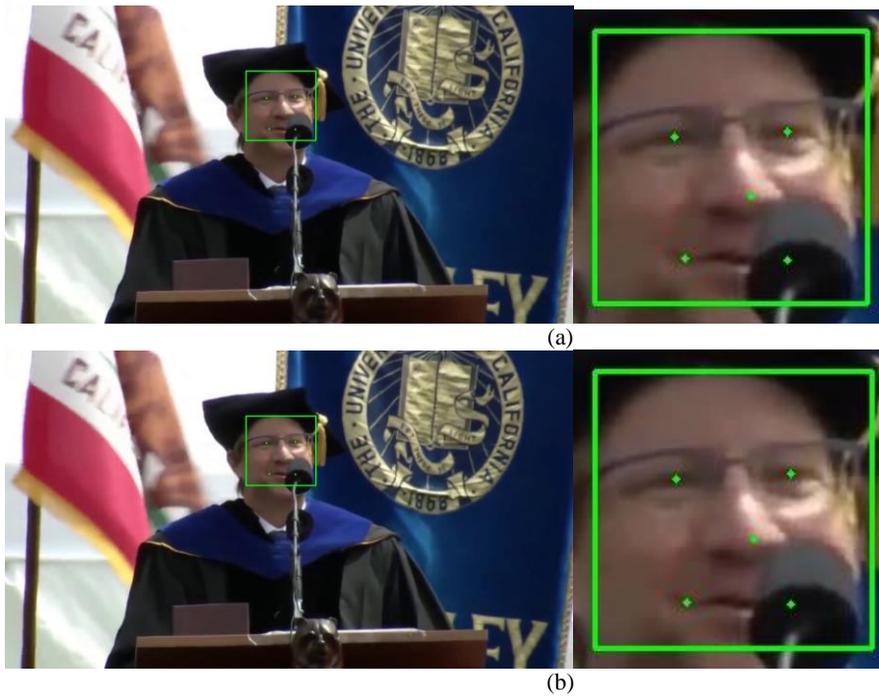

Fig. 2. The facial landmark detection results in the 1st frame achieved by (a) the method combining [5] with [20], (b) the proposed DTD framework

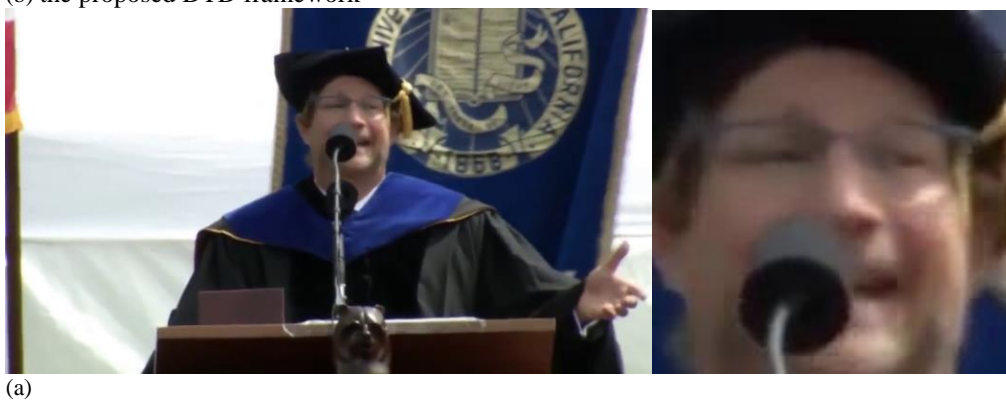

(a)

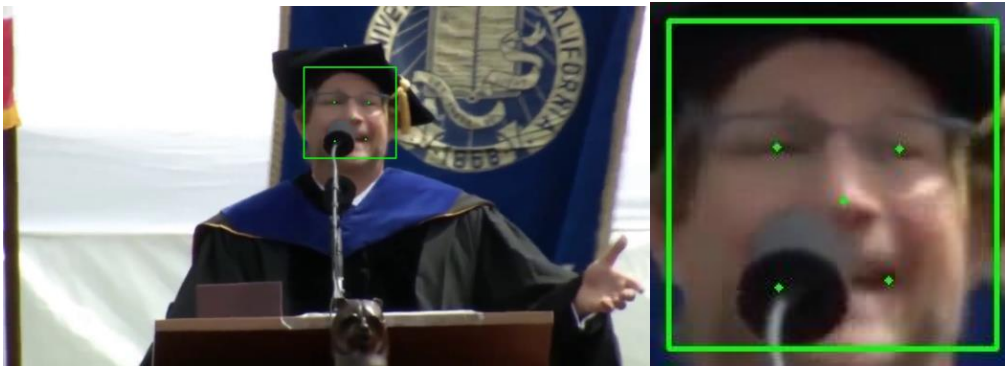

(b)
Fig. 3. The facial landmark detection results in the 682th frame achieved by (a) the method combining [5] with [20], (b) the proposed DTD framework

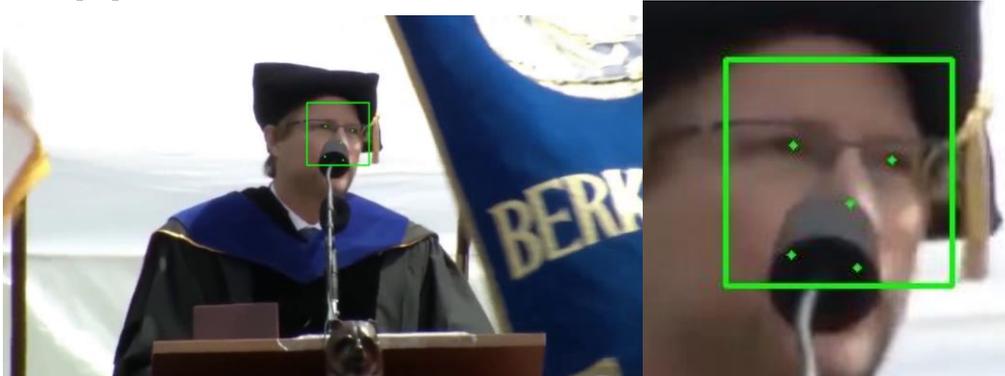

(b)
Fig. 4. The facial landmark detection results in the 1242th frame achieved by (a) the method combining [5] with [20], (b) the proposed DTD framework

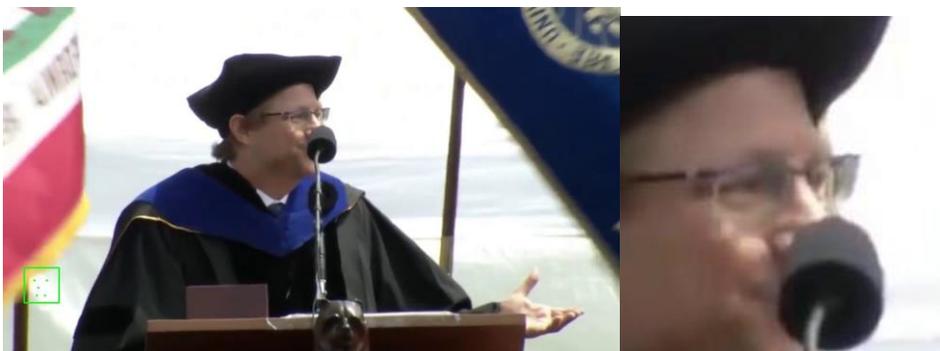

(a)

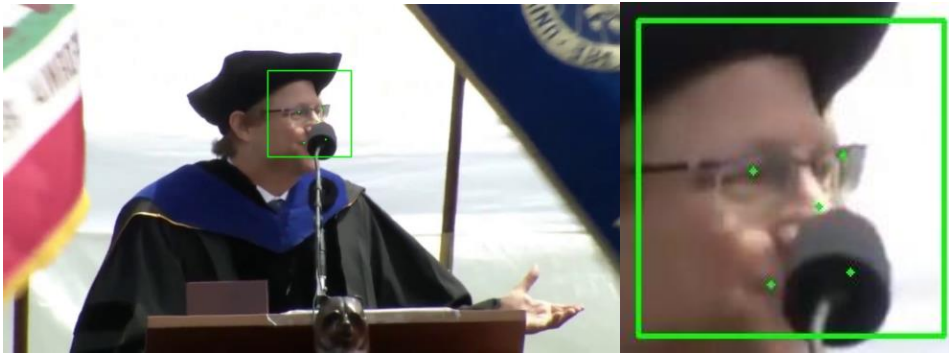
(b)
Fig. 5. The facial landmark detection results in the 1673th frame achieved by (a) the method combining [5] with [20], (b) the proposed DTD framework

## 4 Conclusions

In this paper, we present a novel hybrid framework termed as detection-tracking-detection (DTD) to dynamically detect the facial landmarks in the videos. The hybrid framework involves global/local face detection, facial landmark detection and facial landmark tracking. Experimental results show that (1) our framework is robust since geometric constraints among facial landmarks are implicitly utilized by the deep convolutional neural network (DCNN). This is the first time, to the best of our knowledge, to apply the cascaded DCNN to facial landmark detection in the videos. (2) The face bounding box is effectively returned by incorporating the idea of facial landmark tracking. This scheme also has the real-time advantage. In the future, the performance of the proposed framework will be further improved in the cases of large area occlusions and large pose variations which are also open and challenging problems in the fields of computer vision. We consider that this task could be implemented by adding more challenging samples to train the cascaded DCNN in the proposed framework. Also, we will take more semantic facial landmarks to establish stronger facial geometric constraints in the future.

## 5 Acknowledgements

This work was supported in part by the National Natural Science Foundation of China (Grant Nos. 61001179 and 61571139), Guangdong Science and Technology Plan (Grant Nos. 2015B010104006, 2015B010124001 and 2015B090903017).